# Research on geometric figure classification algorithm based on Deep Learning

Ruiyang Wang, Haonan Wang, Junfeng Sun, Mingjia Zhao, Meng Liu

School of Sciences, Liaoning Technical University, Fuxin, Liaoning, 123000, China

## Abstract

In recent years, with the rapid development of computer information technology, the development of artificial intelligence has been accelerating. The traditional geometry recognition technology is relatively backward and the recognition rate is low. In the face of massive information database, the traditional algorithm model inevitably has the problems of low recognition accuracy and poor performance. Deep learning theory has gradually become a very important part of machine learning. The implementation of convolutional neural network (CNN) reduces the difficulty of graphics generation algorithm. In this paper, using the advantages of lenet-5 architecture sharing weights and feature extraction and classification, the proposed geometric pattern recognition algorithm model is faster in the training data set. By constructing the shared feature parameters of the algorithm model, the cross-entropy loss function is used in the recognition process to improve the generalization of the model and improve the average recognition accuracy of the test data set.

## 1. Introduction

With the development of big data era and modern artificial intelligence technology, deep learning of graphic recognition has been applied in various fields of modern society[1]。 The traditional computer geometric figure recognition algorithm technology can be divided into two categories, the first is non interactive computer graphics, the other is interactive computer graphics. However, both of them mainly focus on image recognition, and need to preprocess the image artificially in the early stage and classify the image based on manual experience. Artificial intelligence developed rapidly around 2016,Deep learning theory has made achievements in many fields, such as natural language processing, speech recognition and so on[2]People begin to train and simulate a large number of data models with the help of deep learning.In this paper, a geometry classification algorithm model based on convolution neural network in deep learning is established. In the process of graph classification, the cross entropy function is used to enhance the generalization of the algorithm model. Finally, 300 geometric figures classification experiments are completed.

## 2. Deep learning theory

### 2.1. Deep learning theory

Machine learning is a special subject that studies how computers simulate or realize human learning behavior to acquire new knowledge. The deep learning theory is one of unsupervised learning in machine learning. Its core idea is to simulate the neural network of human brain learning to interpret data[3]Compared with the traditional shallow learning, deep learning emphasizes the depth of the algorithm model structure. Through the layer by layer feature transformation of the data set, the feature representation of the sample in the original space is transformed into a new feature space, so as to improve the classification and prediction ability of experimental results.The typical modeling methods in deep learning include convolution neural network and recursive neural network.

## 2.2. Convolution neural network theory

Convolution neural network structure belongs to artificial neural network, which is mainly composed of five parts, which is composed of two-layer network association structure of convolution layer and correlation sampling layer. The upper and lower levels of the set rules are connected with each other, and the adjacent neural organizations are interconnected, so as to form the reinforcement between levels[4]。 The network structure diagram is divided into input layer and hidden layer. The hidden layer refines the convolution layer and pooling layer again. Each neuron is connected with each other to construct a supervised learning training network[5]The training methods of the network model are divided into the former training propagation and the reverse training propagation, as shown in Figure 1 and Figure 2:

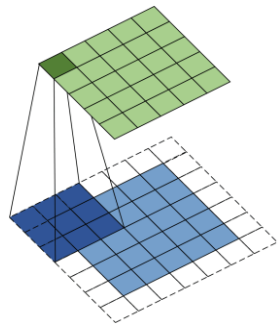

Figure 1 forward propagation process

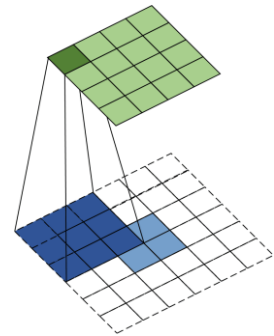

Figure 2 back propagation process

Among them, the convolution kernel in the mapping feature graph and convolution neural network is calculated, and the two are combined and input each other. The algorithm formula is as follows:

$$s(i,j) = \sum_{a=0}^{A}\sum_{b=0}^{B}(w_{a,b}, x_{i+a} + w_{m})$$

It uses $s(i, j)$ Construct eigenvector matrix, and then use $A \times B$ Label the two-dimensional input matrix, $w_{a,b}$ Represents the convolution kernel matrix, $w_m$ Represents the offset.The cross entropy method is used to calculate the loss function used in the construction of the network model

$$L = \frac{1}{n}\sum_{x}[y\ln(a) + (1-y)\ln(1-a)]$$

Where l is the loss function parameter, X is the input sample data, a is the calculation result, y is the label sheet, and N is the total number of samples. The general process of convolution neural network model is shown in Figure 3

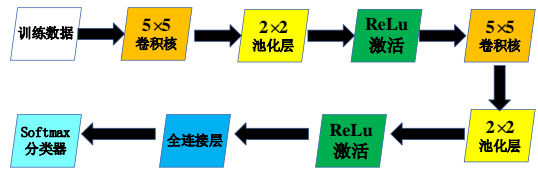

Fig. 3 approximate process of convolutional neural network

## 3. Model design based on depth CNN

In order to get a better recognition result of geometry training model, we need to reconstruct the deep CNN model. In this paper, we use the framework of CNN's classic model lenet-5 to establish the algorithm model of geometric pattern recognition[6]The specific algorithm model flow is as follows:

*Step 1:* build geometry database.The geometry test set database is derived from the geometry of different shapes in kaggle website, and then the data expansion operation is carried out to obtain the geometry training set database.

*Step 2:* set the lenet-5 architecture model.

The lenet-5 architecture model combines feature extraction, image recognition and self-learning, continuously calculates the weight matrix of convolution kernel through forward propagation, and then the network carries out back-propagation to change the local connection and weight sharing matrix, which is composed of one input layer, two convolution layers, two pooling layers and two full connection layers. The specific architecture is shown in Figure 4

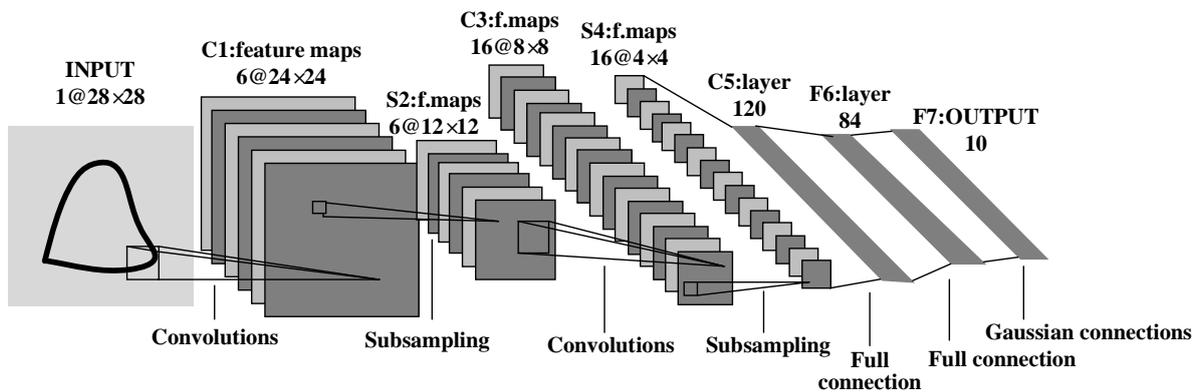

Figure 4 Schematic diagram of lenet5 architecture

*Step 3:* set the parameters of convolution layer. Convolution layer C1 input image, through a step size of 2 5×5 convolution kernel matrix, reduce the image size, and featuer maps is 6

$$con = f(\sum_{i,j \in M} x_{i,j} \Lambda W_{m-i,n-i} + b)$$

Where X represents the elements to be input into m area of the architecture model, I and j are the location coordinates, W is the convolution kernel element, m and N are the sizes of the convolution kernel, and B is the offset. $f(\Lambda)$ Is the softmod classifier function.

*Step4:* set pooling layer parameters. The feature map of the pooling layer S2 turns the image into the maximum pooling layer. The size of the feature map is halved to obtain the maximum pooling. The calculation process of the pooling layer is as follows:

$$px = A(\max(y_{i,j})),$$

Among them, Y represents the elements in the pooling area p, I and j belong to the element coordinates, a represents the down sampling, the convolution layer C3 and pooling layer S4 repeat the above operation, and the convolution kernel step size is 1 and the size is 5 after the final feature maps are obtained×Five convolution operation, get two 1×120.

*Step 5:* use the training set to carry out three types of graphics on the 300 test set database, classify them as triangle, circle and square, and generate the geometric pattern recognition results of the algorithm.

## 4. Result analysis of figure recognition

The experimental platform used in this paper is python38, PyCharm。 The data of the test set is from the kaggle website. There are 300 geometric figures of different shapes. The images are 28 * 28 gray images, and the main graphic contents are triangles and circles. The test set images are shown in Figure 5.

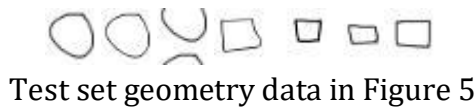

Test set geometry data in Figure 5

Due to the small size of the picture and the unclear content, it may bring some challenges to the classification. Therefore, a series of operations such as turning, folding and translating are randomly performed on the graph in each test set database to enhance the diversity of data feature extraction and generate a training set database of 2100 different geometric graphs.

Firstly, the lenet-5 network model is trained with 10 training times, 0.001 learning rate and 0.9 momentum test parameters.Firstly, after lenet-5 architecture training, different convolution kernels can continue to change back-propagation parameters after forward propagation calculation. The schematic diagram of back-propagation parameters is shown in Figure 6:

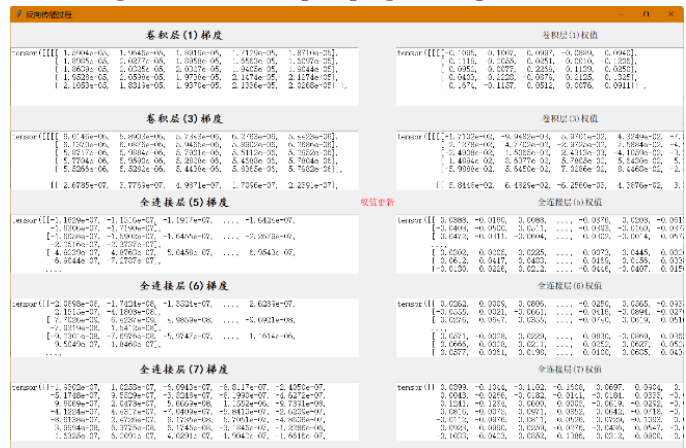

Fig. 6 parameter diagram of back propagation process

As shown in Figure 7, you can see the final set graph classification result of any geometric graph imported into the test set, as shown in Figure 8.

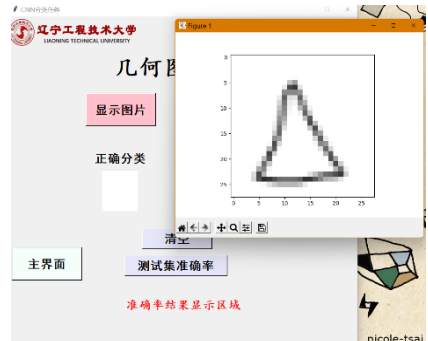

Figure 7 Random image import process

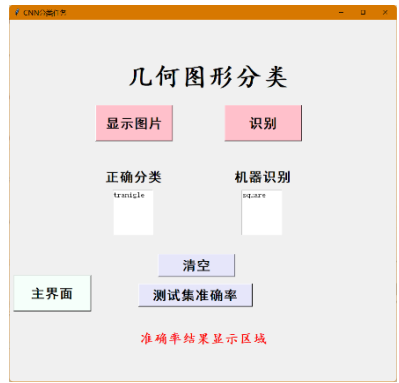

Fig. 8 Final image classification results

According to the experimental results of network training times and learning rate before and after, the final pattern diagram of training set loss error can be obtained, as shown in Figure *:

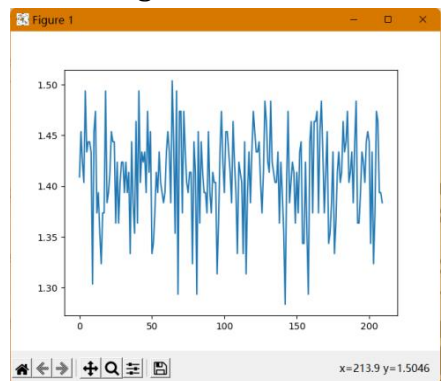

Fig. 9 Model residual results

It is found that the size of different geometric figures has great influence on the accuracy rate. After a small amount of figures are input in the unified network, the classification rate can be higher, but it takes more training time. However, the final network training model can classify the geometric figures in the test set in different ways, such as triangle, circle and square. The final result can be obtained, and the final accuracy rate can reach 90%.

## 5. Summary

The deep learning CNN algorithm has also made remarkable achievements, making image classification and recognition intelligent, greatly improving the actual accuracy of image recognition. In this paper, a classification model of geometric figures is established with the help of the classical model LeNet-5 of CNN in deep learning. In the model training stage, a model network is built to share characteristic parameters.In the training process, the advantages of convolutional neural network, such as sharing weights, self-learning to extract classification features and network training, are combined. In the classification process, the cross entropy loss function is used to improve the generalization and accuracy of the model. However, as far as the present situation is concerned, there are still many problems, such as complicated image processing and long preprocessing time. In the process of introducing GPU,Because the data interface is not universal, it is necessary to select and input relevant data sets.